\documentclass[runningheads]{llncs}
\usepackage{graphicx}
\usepackage{caption}
\usepackage{subcaption}
\usepackage{cite}
\usepackage{amsmath,amssymb,amsfonts}
\usepackage{amsmath}
\usepackage{esint}
\usepackage{tabularx}
\usepackage{amssymb}
\usepackage{mathrsfs}
\usepackage{textcomp}
\usepackage{xcolor}
\usepackage{algorithm}
\usepackage[noend]{algpseudocode}
\def\BibTeX{{\rm B\kern-.05em{\sc i\kern-.025em b}\kern-.08em
    T\kern-.1667em\lower.7ex\hbox{E}\kern-.125emX}}
\begin{document}
\title{Decentralized federated learning methods for reducing communication cost and energy consumption in UAV networks}
\titlerunning{DFL methods for reducing Comms Cost and E Cost in UAV-N}
%
\author{Deng Pan\inst{1} \and
Mohammad Ali Khoshkholghi\inst{2} \and
Toktam Mahmoodi\inst{3}}
\authorrunning{Deng Pan et al.}
%
\institute{University College London, United Kingdom\\ 
\email{deng.pan.22@ucl.ac.uk} \\
King's College London, London, United Kingdom\\
\email{ali.khoshkholghi@kcl.ac.uk}\\
King's College London, London, United Kingdom\\
\email{toktam.mahmoodi@kcl.ac.uk}}
\maketitle              
\begin{abstract}
Unmanned aerial vehicles (UAV) or drones play many roles in a modern smart city such as the delivery of goods, mapping real-time road traffic and monitoring pollution. The ability of drones to perform these functions often requires the support of machine learning technology. However, traditional machine learning models for drones encounter data privacy problems, communication costs and energy limitations. Federated Learning, an emerging distributed machine learning approach, is an excellent solution to address these issues. Federated learning (FL) allows drones to train local models without transmitting raw data. However, existing FL requires a central server to aggregate the trained model parameters of the UAV. A failure of the central server can significantly impact the overall training. In this paper, we propose two aggregation methods: Commutative FL and Alternate FL, based on the existing architecture of decentralised Federated Learning for UAV Networks (DFL-UN) by adding a unique aggregation method of decentralised FL. Those two methods can effectively control energy consumption and communication cost by controlling the number of local training epochs, local communication, and global communication. The simulation results of the proposed training methods are also presented to verify the feasibility and efficiency of the architecture compared with two benchmark methods (e.g. standard machine learning training and standard single aggregation server training). The simulation results show that the proposed methods outperform the benchmark methods in terms of operational stability, energy consumption and communication cost.

\keywords{Federated Learning  \and Unmanned Aerial vehicles \and Decentralized Training.}
\end{abstract}
\section{Introduction}
Unmanned aerial vehicles (or drones) will positively impact society by supporting various services for the modern smart city. Examples include applications in goods delivery, real-time road traffic monitoring, target identification, etc. \cite{1}. Machine learning (ML) to give UAVs network intelligence is a crucial requirement to enable such applications. However, traditional machine learning techniques require uploading all data to a cloud-based server for training and processing, which represents a considerable challenge for drone swarms \cite{2}.

In a first consideration, the data generated by drones may be sensitive, and could be intercepted while uploading the data to the cloud, leading to a privacy breach. Secondly, drones' large numbers of data can result in impractical delays when uploading, thus creating a time lag for swarms of drones  that prevents them from conducting real-time monitoring. Finally, drones can consume a great amount of energy when training models, meaning there may be related challenges to doing so in terms of energy constraints \cite{3}.

Distributed machine learning techniques represent a new solution to address these issues and challenges, whereby drones train machine learning models without sharing raw data. Federated learning (FL), recently developed and proposed by Google as an emerging distributed machine learning technology, will provide further new technology to support the intelligence of drones \cite{4}. The concept of federated learning is allowing each drone to train its learning model based on its data. The parameters of each drone's trained model are then sent to a parameter server to update the model for a new round of training, without sending the raw data to the cloud. This training model allows for reasonable data security, latency and energy consumption.

However, the highly mobile nature of drones means conventional FL is not well-suited, given their complex working environment. If the parameter server does not work properly, it will impact the training effectiveness of the whole UAV network \cite{5}. At the same time, each drone needs to return to the parameter server after each local training session to upload parameter information; and then return to where it is supposed to work after updating the parameters, which increases the working time and wear and tear on the drone's components. Also, if a large number of UAVs access the parameter server simultaneously to upload parameter information and update the local model, it is a test of the bandwidth of the parameter server \cite{6}. In this context, a fully decentralised federated learning architecture holds the potential to significantly optimise existing intelligent drone networks.

To address the current problem, Decentralized Federated Learning for UAV Networks (DFL-UN), proposed by Qu et al., proposes establishing links between UAVs in a small area while setting any UAV in range to act as a simulation server to aggregate models \cite{7}. The authors address FL's centralisation and user selection issues for UAV network but do not account for UAVs' limited power supply. The main objective of this work is to optimise the architecture of the DFL-UN while guaranteeing training results and analysing communication cost and energy consumption. As our contributions in this work, first, classify UAVs in the region into two groups to determine which two UAVs will be used as aggregation centres. Then, we optimise the decentralised algorithm proposed by Liu et al. to derive two aggregation methods, Commutative FL and Alternate FL \cite{8}. The method we proposed is compared with cases where FL is not used, with only one aggregation centre. The simulation results validate the convergence of commutative FL and alternate FL in the architecture, demonstrating more stable training results than when not using FL and with only one aggregation centre. 

The rest of this paper is outlined below. We will briefly introduce the relevant research literature in Section 2. Then we will briefly describe the design and implementation of this work in Section 3. Nest, we will discuss the collected and listed research findings in Section 4. The conclusion will be given in the final section 5.

\section{Background and Related Work}

\subsection{Definition of Federated Learning}

\begin{figure}
	\includegraphics[width=0.6\textwidth]{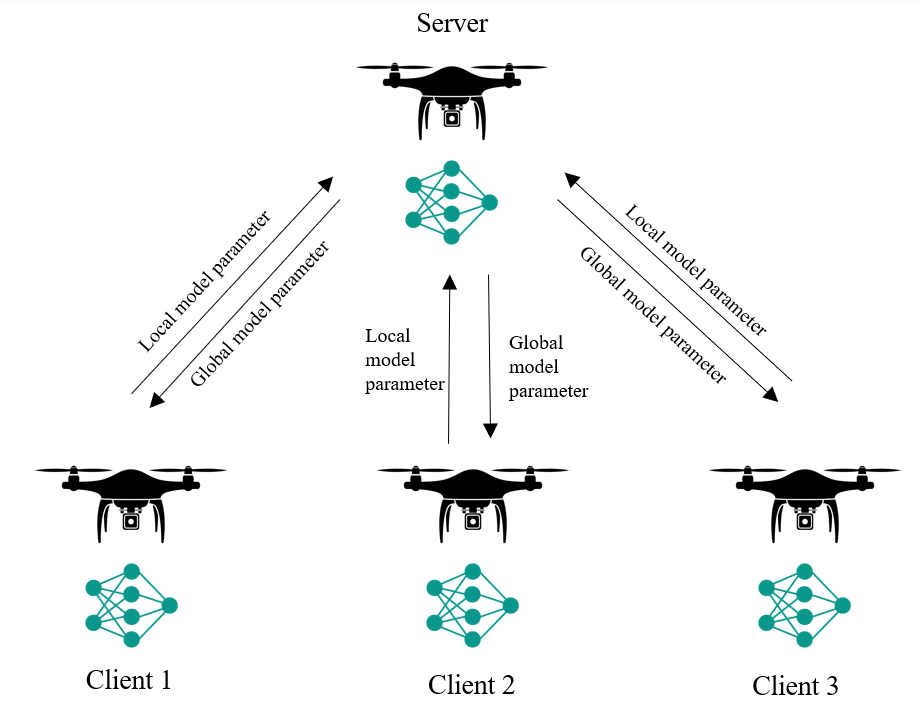}
    \centering
    \caption{FL framework}
    \label{fig.1}
\end{figure}

The term federated learning (FL) was first coined by McMahan et al. in 2016, who defined it as ``a learning task that is solved by a loosely federated participating device (client) that is coordinated by a central server"\cite{4}. We can now define FL more broadly by adopting this initial definition: FL is a privacy-preserving distributed machine learning technique where N participants (clients) train the same model using their locally stored data. Eventually, producing a federated model on a central server by exchanging and aggregating the model parameters for each client and using the federated model to update the model on the client's side \cite{9, 10}. The framework of FL is shown in Figure \ref{fig.1}.

\subsection{Overview of the Federated Average Algorithm (FedAvg)}

The literature \cite{4} proposes using the federated averaging algorithm (FedAvg) to train models in federated learning. Assuming that the client has the initial model, in round $t$ when the central model parameters are updated, the $k$-th participant will compute the batch gradient $g_{k}$, and the server will aggregate these gradients and use the updated information on the model parameters according to the following formula.
\[w_{t+1} \leftarrow w_{t}-\eta \sum_{k=1}^{K} \frac{n_{k}}{n} g_{k} \tag{2-3}\]
where $\eta$ is the learning rate and $\sum_{k=1}^{K} \frac{n_{k}}{n} g_{k} = \Delta f(w_{t})$. FedAvg is the average gradient $w$ sent to each participant, who will calculate the updated model parameters w according to Equation 2-3.
\[\forall k, w_{t+1}^{k} \leftarrow w_{t}-\eta g_{k} \tag{2-4}\]
\[w_{t+1} \leftarrow \sum_{k=1}^{K} \frac{n_{k}}{n} w_{t+1}^{k} \tag{2-5}\]
Each client performs gradient descent locally on the existing model parameters $w_{t}$ using local data according to Equation 2-4 and sends the locally updated model parameters $w_{t+1}^{(k)}$to the server. The server then computes a weighted average of the model results according to Equation 2-5 and sends the aggregated model parameters $w_{t+1}$ to each participant. The complete pseudo-code of FedAvg is as follows
\begin{algorithm}
	\caption{FederatedAveraging. $K$ clients are indexed by $k$; $C$ is the  percentage of clients performing the computation in each round, $B$ is the local minibatch size, $E$ is the number of local epochs, $\mathcal{P}_{k}$ denotes the index set located on the data side of participant $k$ and $\eta$ is the learning rate\cite{4}}
	\textbf{Server executes}:
	\begin{algorithmic}[1]
        \State initialize $w_{0}$
		\For {each round $t=1,2, \ldots$}
        \State $m \leftarrow \max (C \cdot K, 1)$
        \State $S_{t} \leftarrow$ (random set of $m$ clients)
        \For {each client $k \in S_{t}$ \textbf{in parallel}} 
        \State  $w_{t+1}^{k} \leftarrow$ ClientUpdate $\left(k, w_{t}\right)$ 
		\EndFor
		\State \textbf{end for}
		\State $w_{t+1} \leftarrow \sum_{k=1}^{K} \frac{n_{k}}{n} w_{t+1}^{k}$
		\EndFor
		\State \textbf{end for}
	\end{algorithmic}
	\textbf{ClientUpdate($k$, $w$):} //\textit{Run on client k}
	\begin{algorithmic}[1]
        \State $\mathcal{B} \leftarrow \text { (split } \mathcal{P}_{k} \text { into batches of size } B \text { ) }$
		\For {each local epoch $i$ from 1 to $E$}
        \For {each batch $b\in \mathcal{B}$} 
        \State  $w \leftarrow w-\eta \nabla \ell(w ; b)$ 
		\EndFor
		\State \textbf{end for}
		\EndFor
		\State \textbf{end for}
	\end{algorithmic}
\end{algorithm} 
\subsection{Decentralised Federated Learning for UAV Networks}
As stated in the literature \cite{7}, FL training may be aborted due to the mobility of UAV. DFL-UN represents a solution to use in such situations.
The prominent architecture of DFL-UN is based on a fully distributed scenario without a central server or a fixed UAV as a parameter server. Each drone is trained with local data, and neighbouring drones receive the model parameters. The DFL-UN architecture operates in the following steps.\par
    -- All drones are pre-installed with the FL training model. A built-in coordinator is responsible for distributing central information to all designed drones and monitoring the FL training process.
    -- Each drone will train a local model parameter in training round $t$.\par
    -- Firstly, in round $t$, drones $i+1$, $i+2$ and $i+3$ send their training models to $W_{i+1,t}$, $W_{i+2,t}$ and $W_{i+3,t}$ to drone $i$, where $W_{i,t}$ represents drone $i$'s local model parameter at training round $t$.\par
    -- Secondly, drone $i$ aggregates $W_{i+1,t}$, $W_{i+2,t}$,  $W_{i+3,t}$ and including drone $i$'s training model parameter to generate an aggregated model parameter.\par
    -- Thirdly, drone $i$ will ``broadcast'' these aggregated model parameters to its neighbouring drones for model updates, and drone $i$ will also update its local model.

\subsection{Decentralised Federated Learning (DFL)}
An algorithm that alternates between local updates and inter-node communications is proposed in the literature \cite{8} to cope with the absence of a central server in DFL. This algorithm is broadly similar to FedAvg but takes a new approach to decentralisation. Liu et al. proposed dividing the communication steps into $\tau_{1}$ and $\tau_{2}$, i.e. the frequency of local updates and the frequency of inter-node communications, respectively \cite{8}. Let $\tau = \tau_{1} + \tau_{2}$ be defined in the DFL framework as an iteration round, such that for the $k$th iteration round, $((k-1)\tau, k\tau)$ for k = 1,2,.... The complete pseudo-code is as follows.
\begin{algorithm}[H]
	\caption{DFL\cite{8}}
	\textbf{Parameters}:\par
	\qquad Learning rate $\eta$, total number of steps $T$, computation frequency $\tau_{1}$ and communication frequency $\tau_{2}$ in an iteration round, $C \in \mathbb{R}^{N \times N}$ is the confusion matrix and $X_{t}$ is the model parameter matrix 
    \end{algorithm}
    \begin{algorithm}                     
    \begin{algorithmic} [1] 
        \State Set the initial value of $X_{0}$
		\For {$t=1,2, \ldots, K_{\tau}$}
        \If {$t \in [k]_{1}$ where $k=1,2, \ldots, K$}
        \State $X_{t+1} = X_{t} - \eta G_{t}$\qquad \text{$local\ update$}
        \Else
        \State  $X_{t+1} = X_{t}C$\qquad  \text{$inter-node\ communication$}
		\EndIf
		\State \textbf{end if}
		\EndFor
		\State \textbf{end for}
		\For {$t=K_{\tau+1}, \ldots,T $}
		\State $X_{t+1} = X_{t} - \eta G_{t}$
		\EndFor
		\State \textbf{end for}
	\end{algorithmic} 
\end{algorithm} 

\section{Design and Implementation}
\subsection{UAV model}
In this work, 10 or 20 drones were randomly generated in a $10 \times 10$ $m^{2}$ area. The drones are then divided into two clusters using the K-means clustering algorithm, and the coordinate information of each drone is recorded by themselves \cite{11}. The default drone had enough battery to support the drone for 1 hour of flight and operation. Next, each drone was assigned a model and local data to be used for training. Finally, the coordinates, client ID and remaining battery level (first recorded as 100\%) of all drones in the current training round were recorded for subsequent data analysis.
\subsection{Aggregation Model}
For training that required FL, we determined whether the cluster head was single or double, so the appropriate training function could be selected. The training functions take different patterns depending on the training method, of which there were four, based on different UAV classifications, in this work:
\begin{itemize}
\item Commutative FL ($C$). This model is the case of classifying the UAV into two clusters, where the client and aggregate server within each cluster perform $n$ intra-cluster FL and $m$ inter-cluster exchanges of FL.\\
\item Alternate FL ($A$). This model is the case where the drone is divided into two clusters, and the subsequent training round after each intra-cluster FL will be the inter-cluster swapped FL.\\
\item One-server FL ($One$). This is one of the control group. In this model, the UAV is grouped into one cluster, similar to the FL in the standard case.\\
\item Normal Machine Learning ($O$). This is another one of the control group. In this model, each drone will only be trained with local data, and no data exchange occurs. 
\end{itemize}
\begin{figure}[!h]
    \centering
     \begin{subfigure}[b]{0.45\textwidth}
         \centering
         \includegraphics[width=1\textwidth]{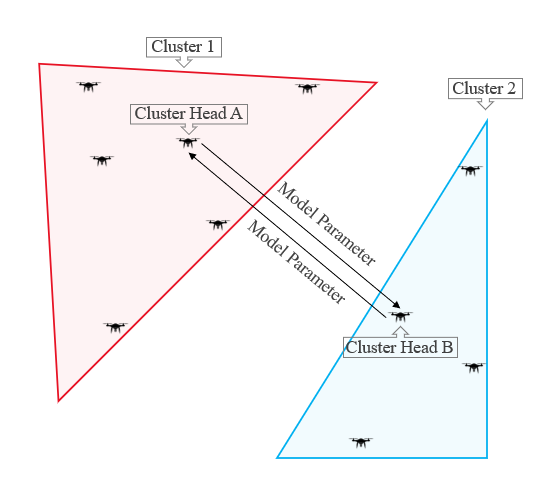}
     \end{subfigure}
     \hfill
     \begin{subfigure}[b]{0.45\textwidth}
         \centering
         \includegraphics[width=1\textwidth]{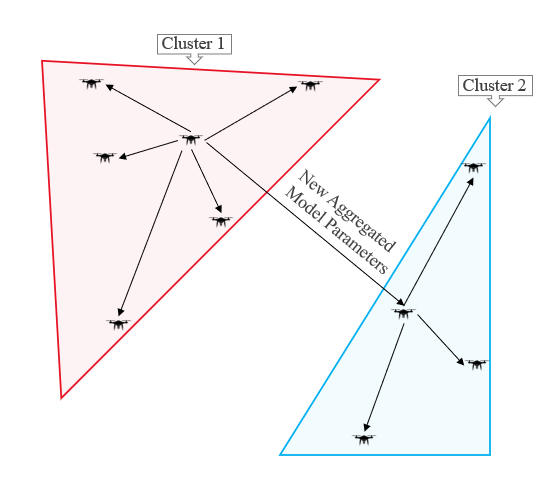}
     \end{subfigure}
     \caption{Inter-cluster FL framework}
     \label{fig.2}
\end{figure}
The inter-cluster FL exchanged between the clusters mentioned above refers to an additional round of aggregation and evaluation computation by two clusters uploading each other's updated model parameters after the FL ends within each the cluster. Suppose the two cluster heads are A and B, and aggregated model performance sent by B as the client to A as the server is better than that sent by A as the client to B as the server. All drones in both clusters will receive the new aggregated model parameters sent by B to A and vice versa. The framework of inter-cluster FL is shown in Figure \ref{fig.2}. The process of training and updating the drone information is repeated until a valid model with stable accuracy and loss values is obtained. After this, the accuracy, loss, battery remaining, data sent, data received and total data sent and received for each round were evaluated.
\subsection{Energy Consumption Model}
\paragraph{Local Computation} The drone will perform a deep learning strategy for image recognition locally, and this work calls on the GPU to train it while performing the simulation. For this, we will fix the energy stored in each drone battery as $E_{d}$ and then record the average power of the GPU during the training call to the GPU, $P_{avg}$. The training time, $t_{tr}$ for each drone will then be recorded, and the estimated computational energy consumption will be calculated using $E_{c} = P_{avg}t_{tr}$. It was eventually converted to percentage units by $\text{Batter used} = E_{c}/E_{d}$.
\paragraph{Communication Energy} In this work, all drones are assumed that the wireless channel between each other is connected by Line of Sight (LOS) wireless link. Similar to  \cite{12}, the communication between drone $k$ and another drone is represented as follows:
\[E_{k}^{\mathrm{C}}=t_{k} p_{k} \]
where $t_{k}$ is the time duration to transmit data of size $s$ and $p_{k}$ is the average transmit power of drone $k$. In this work, $t_{k}$ will be calculated according to the theoretical minimum time, which is represented as follows:
\[t_{k}^{\min }=\frac{s}{b_{k} \log _{2}\left(1+\frac{g_{k} p_{k}}{N_{0} b_{k}}\right)}\]
where $b_{k}$ is the bandwidth allocated to drone $k$, $g_{k}$ is the channel gain between two drones and $N_{0}$ is the power spectral density of the Gaussian noise. As in  \cite{13}, we denote the coordinates of UAV $k$ and UAV $k+1$ as $q = \{x, y, z\}$ and $q' = \{x', y', z'\}$, respectively. The channel gain $g_{k}$ of drone $k$ can be calculated by $g_{k}=\hat{\beta}_{0}\left(d_{k} / d_{0}\right)^{-\alpha}$, where $\hat{\beta}_{0}$ is the reference channel gain at $d_{0}=1$ m, $d_{k}$ is the distance between the two drones, and $\alpha$ is the path loss parameter. Hence, the 3D-coordinate distance $d_{k}$ is given as $d_{k} = \sqrt{||q - q'||}$. It is important to note the size of $t_{k}^{min}$ is mainly determined by the distance between the drones and the amount of transmitted data.
\subsection{Communication Cost Model}
In this work, communication costs will be recorded into an logbook file by recording each drone's sent and received file size and the sum of received and sent data. Also, for each method, the communication cost of the drones will be averaged over the file size of all drones received and sent and the sum of received and sent.
\begin{figure}[h]
	\includegraphics[width=0.6\textwidth]{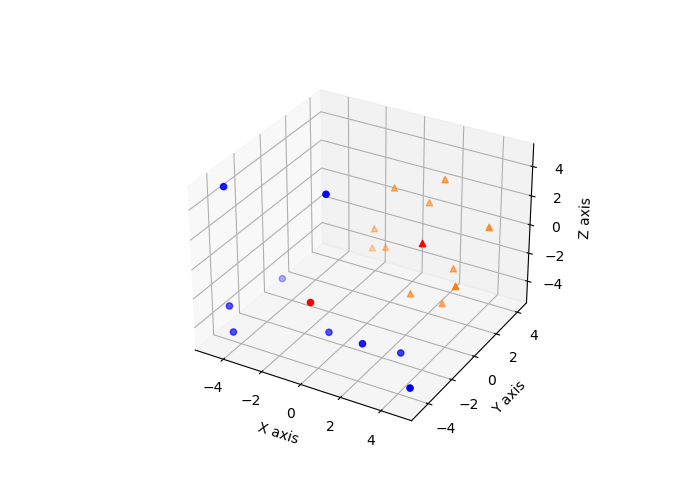}
    \centering
    \caption{20 Drones UAV model: Blue indicates cluster 1, yellow indicates cluster 2 and the red indicate the cluster head}
    \label{fig.3}
\end{figure}
\section{Simulation Result}
In this section, we evaluate the performance of the Commutative FL and Alternate FL and verify the validity of these methods by showing the numerical results of the simulations. The simulated UAV model is first built according to the UAV model in Section 3, as shown in Figure \ref{fig.3}. The datasets used by the UAV for the machine learning model were all taken from the coloured image dataset Cifar-10 \cite{14}. By default, each drone was loaded with a local database of 5000 pieces of information. In addition, the machine learning model used for the UAV uses the deep convolutional network model ResNet-18 \cite{15}. We set the energy possessed by each UAV battery to $E_{d} = 274 $Wh. The bandwidth of each UAV is $b_{k} = 20 $MHz. The reference channel gain is set to $\hat{\beta}_{0} = 28 + 20\log_{10}(f_{c})$ according to the reference path loss specified in \cite{16}, where the carrier frequency $f_{c} = 2 $GHz. The path loss exponent is set to 2.2, the noise power spectral density $N_{0} = -174 $dBm/Hz and the average transmitted power $p_{k} = 10 $dB. For the parameters mentioned above, if not specifically labelled, most parameters are referenced in \cite{12}\cite{13}. After this, to avoid ambiguity, we call the completion of a training ground by the drone a local epoch ($le$) and the completion of a federal learning training round a global epoch ($ge$). The number of federation learning training rounds performed by two clusters in this cluster is called local round ($lr$), and the number of interactions between the two clusters is called global round ($gr$). If not specifically labelled, the default $ge$ will be 30.
\begin{figure}[h]
	\includegraphics[width=0.8\textwidth]{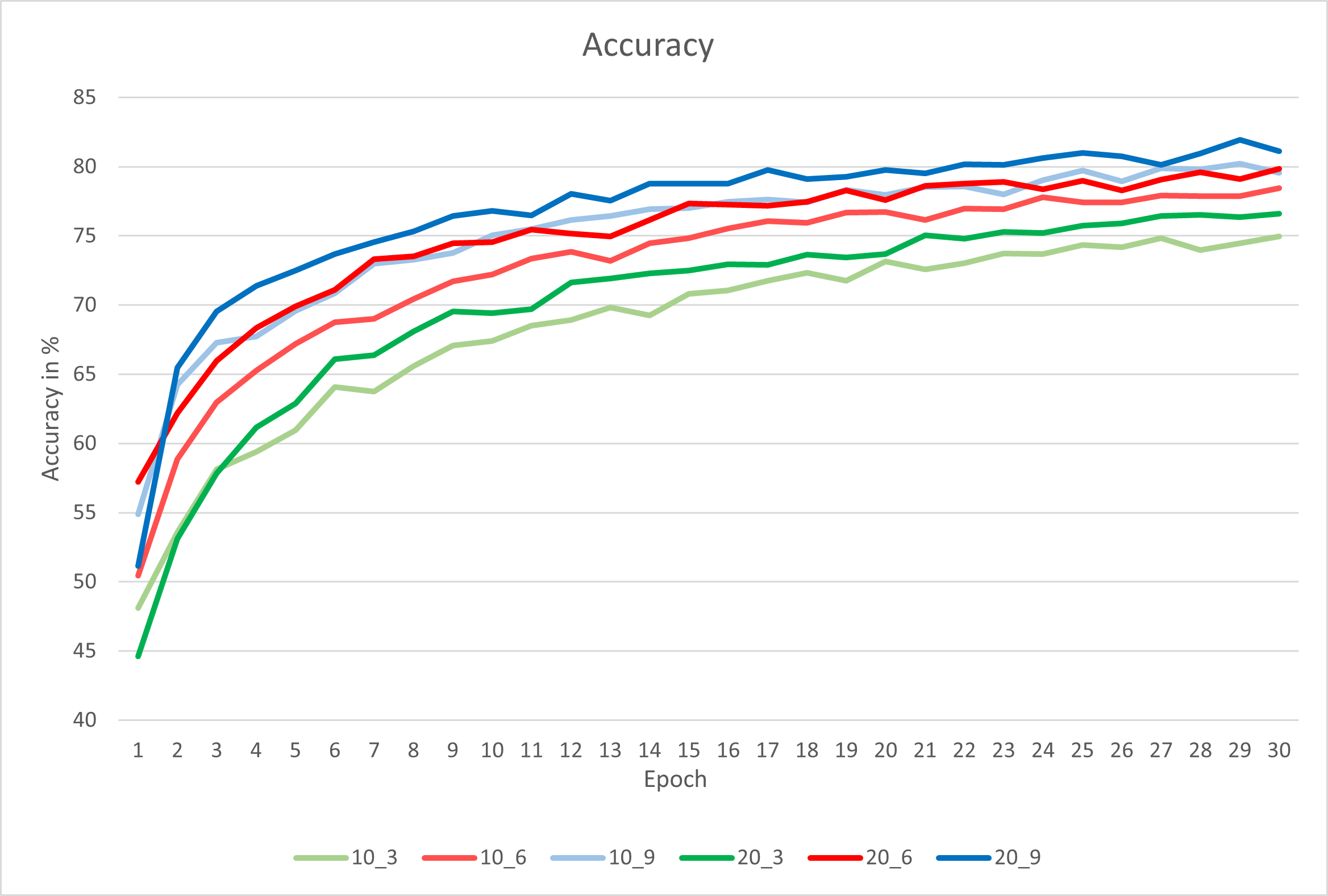}
    \centering
    \caption{Accuracy for training method $C$}
    \label{fig.4}
\end{figure}

Figures \ref{fig.4} shows the model's accuracy for training method $C$, when $lr$ is 5 and $gr$ is 10, and for $le$ values of 3, 6 and 9, respectively. It can be seen from the graphs that 20 drones are generally more accurate than 10 drones when at an equal $le$. This phenomenon was predictable because 20 drones carry twice the amount of data as 10 drones, and therefore, the model should generate this more accurate prediction. We also found that higher $le$ values tended to bring about higher accuracy with the same number of drones. Again, this phenomenon was predictable as more $le$ means faster convergence and more training rounds. At the same time, the accuracy of the two clusters was very close, indicating that the architecture was stable for training method $C$. As shown in Table 1, our simulations show that for training method $C$, different combinations of $lr$ and $gr$ do not have a clear impact on the accuracy of the training but only on the energy consumption and the data amounts sent and received.

\begin{table}[H]
\caption{Data relating to training method $C$} \label{tab1}
\centering
\resizebox{12cm}{!}{
\begin{tabular}{|l|l|l|l|l|l|l|}
\hline
Type & Accuracy & Loss & Avg. Battery & Avg. Send/GB & Avg. Receive/GB & Avg. S\&R/GB \\ \hline
C\_5lr\_5gr\_10  & 75.33 & 0.7242 & 92.00  & 2.47 & 2.47 & 4.94  \\\hline
C\_5lr\_5gr\_20  & 76.68 & 0.6875 & 88.04  & 3.23 & 3.23 & 6.47  \\\hline
C\_5lr\_15gr\_10 & 75.05 & 0.7201 & 93.45  & 2.67 & 2.67 & 5.34  \\\hline
C\_5lr\_15gr\_20 & 77.24 & 0.6642 & 92.49  & 2.90  & 2.90 & 5.79  \\\hline
C\_5lr\_10gr\_10 & 74.98 & 0.7272 & 92.27   & 2.67  & 2.67 & 5.34  \\\hline
C\_5lr\_10gr\_20 & 76.77 & 0.6749 & 92.46  & 2.90  & 2.90 & 5.79  \\\hline
C\_15lr\_5gr\_10 & 74.96 & 0.7293 & 93.07  & 2.07  & 2.07 & 4.15  \\\hline
C\_15lr\_5gr\_20 & 76.78 & 0.6793 & 93.11  & 2.31  & 2.31 & 4.61  \\\hline
C\_10lr\_5gr\_10 & 75.18 & 0.7146 & 93.85  & 2.27  & 2.27 & 4.54  \\\hline
C\_10lr\_5gr\_20 & 77.09 & 0.6743 & 93.15  & 2.50  & 2.50 & 5.01\\ \hline
\end{tabular}
}
\end{table}

\begin{figure}[H]
	\includegraphics[width=0.8\textwidth]{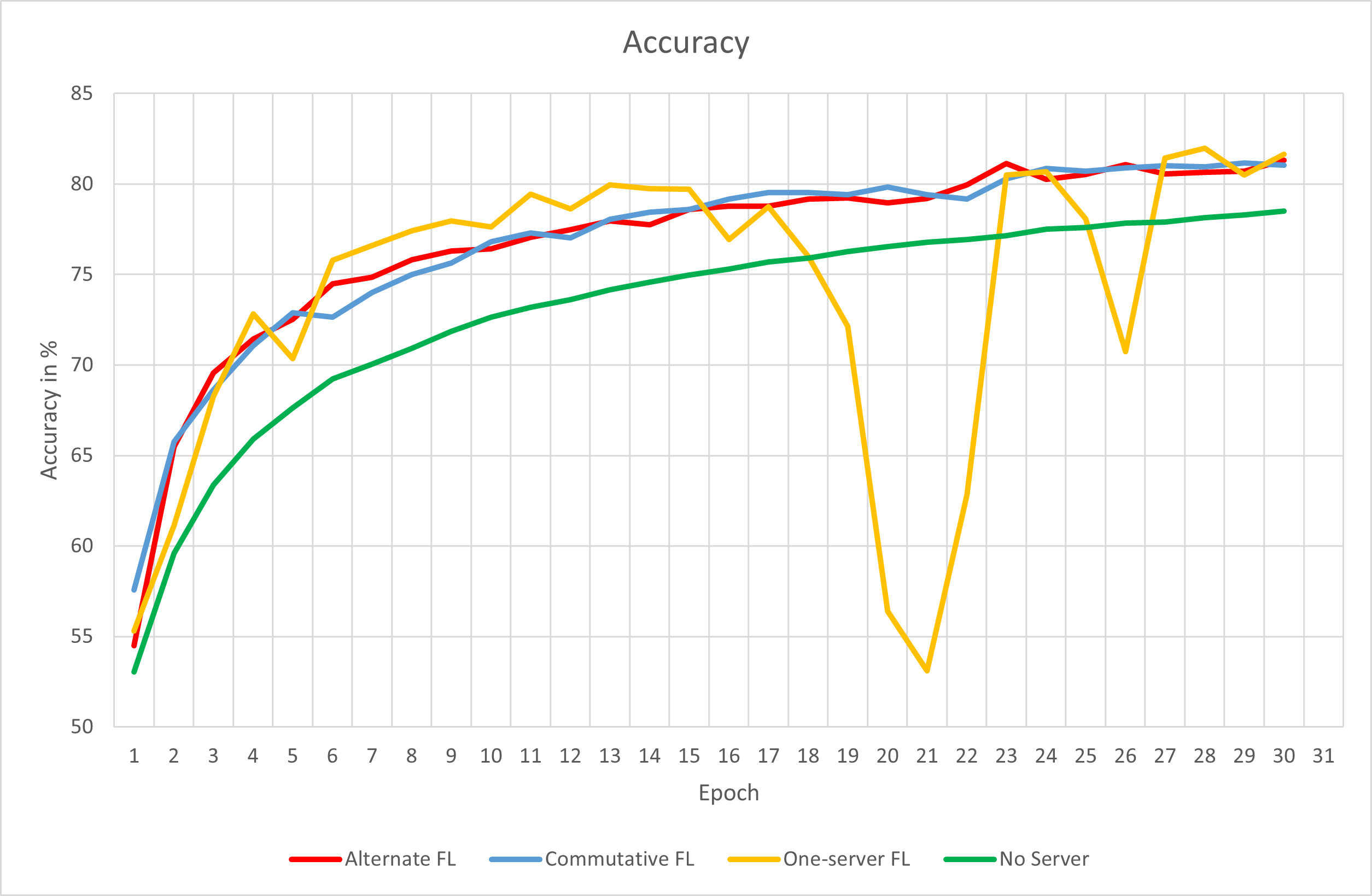}
    \centering
    \caption{Accuracy for training method $C$, $A$, $One$ and $O$}
    \label{fig.5}
\end{figure}

Next, we compare training methods $C$, $A$ and $One$. In the next comparison, training method $C$ will fix the value of $lr$ to 5 and the value of $gr$ to 5. Figure \ref{fig.5} shows how the proposed method compares with other traditional methods in the case of 20 drones. It can be seen that the two proposed methods outperform traditional machine learning methods in terms of accuracy and stability over a single drone federation learning network. The stability of method $One$ with a number of 10 drones is comparable to methods $C$ and $A$. However, it has been tested that for method $One$, the training results fluctuate significantly when the number of drones exceeds 15 in a network. This phenomenon suggests that in the case of method $One$ if the number of drones in the network exceeds 15, the overlap in the data allocated by the drones is higher. This can lead to an overfitted training model, resulting in unstable test accuracy.

\begin{figure}[H]
	\includegraphics[width=0.9\textwidth]{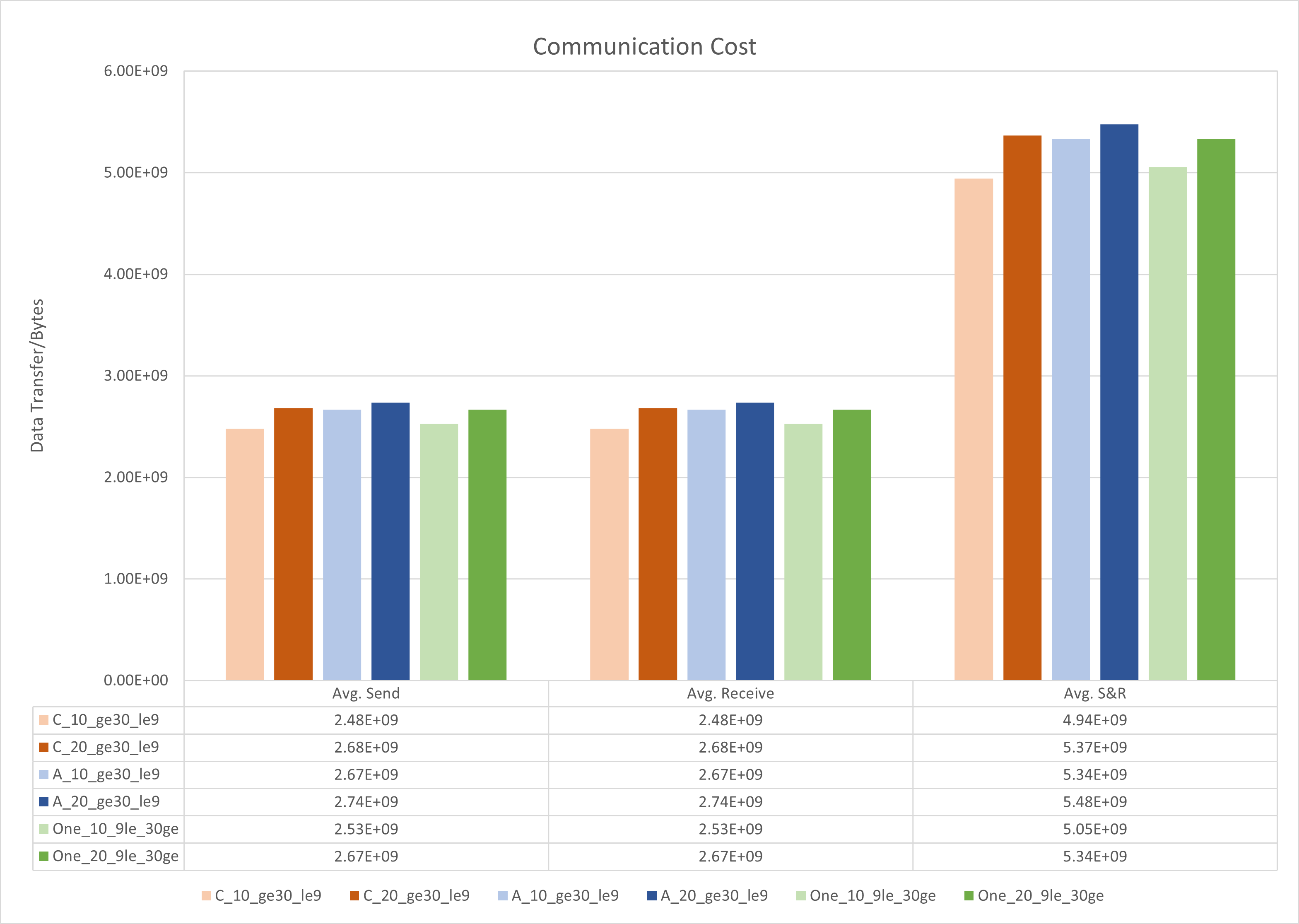}
    \centering
    \caption{Communication Cost for Training Method $C$, $A$ and $One$}
    \label{fig.6}
\end{figure}

\begin{figure}[H]
	\includegraphics[width=0.8\textwidth]{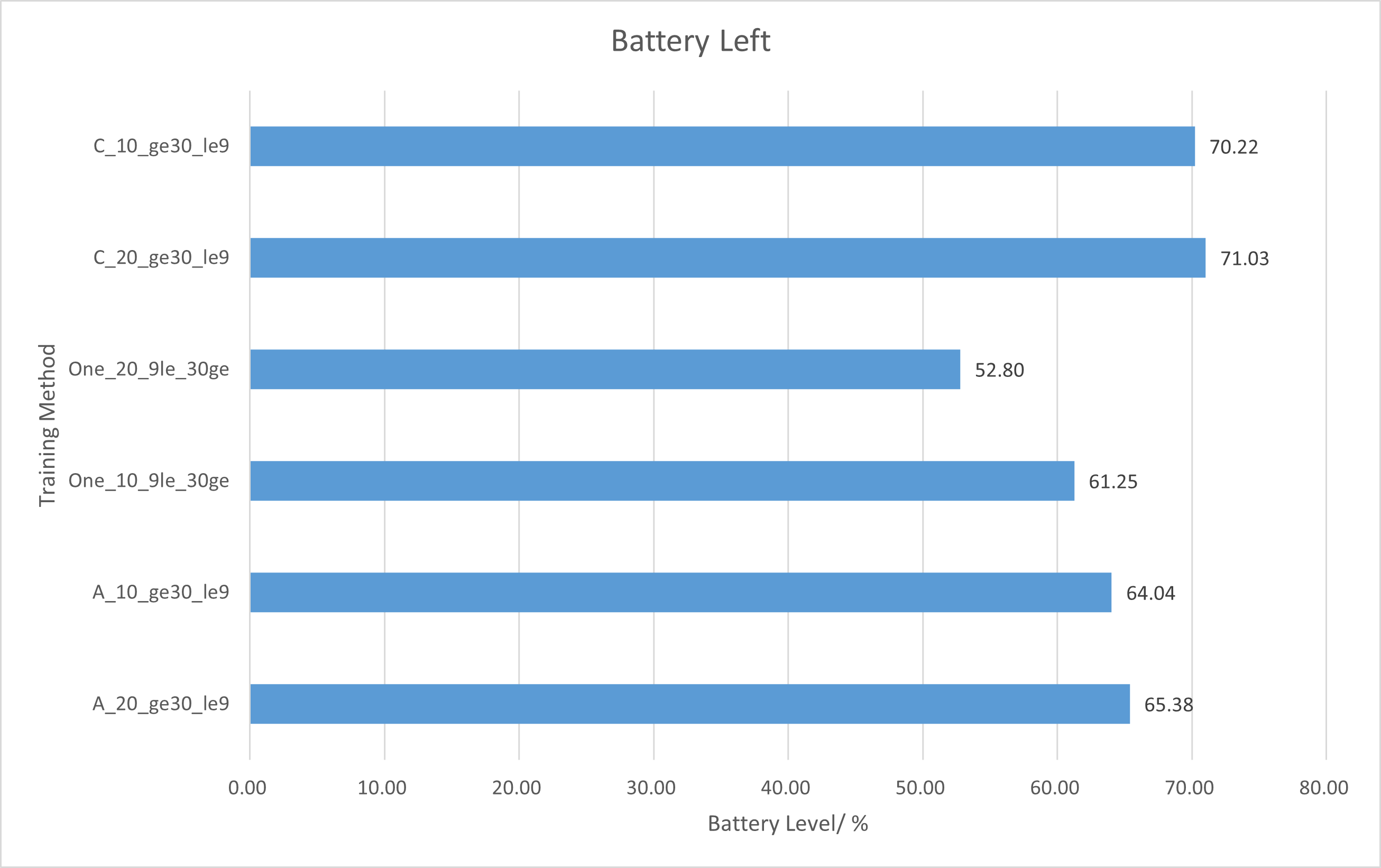}
    \centering
    \caption{Battery Left for Training Method $C$, $A$ and $One$}
    \label{fig.7}
\end{figure}

Figure \ref{fig.6} shows that $C$ outperforms both $A$ and $One$ for a network of 10 drones. The advantage of $C$ over $A$ and $One$ is that $C$ can adjust $lr$ and $gr$ to ensure the training effect while adjusting the amount of data transfer according to the performance of the UAV. On the other hand, Method $A$ fixes the amount of data generated in each training cycle by fixing $lr$ and $gr$, so that the UAV can be assigned a performance to match this value. As seen in Figure \ref{fig.7}, Method $C$ has a higher remaining battery at the end of training than Method $One$ and Method $A$. Methods $C$, and $A$ are very similar in terms of the amount of battery remaining at 10 and 20 UAVs. On the other hand, Method $One$ has a much lower battery left than Methods $C$ and $A$ due to the much higher amount of data handled by a single drone as a parameter server, especially at 20 drones. Again, those figures show the proposed method outperforms conventional FL and ML regarding communication consumption, battery left and training stability.

\section{Conclusion}
In this paper, we investigate the problem of training methods for decentralised federated learning UAV networks. We optimise and propose two learning methods based on existing decentralised federated learning networks to cope with UAVs' communication cost and energy consumption. Simulated numerical results show that our proposed new learning methods can effectively guarantee training results while outperforming conventional training methods in terms of the training stability, communication cost and energy consumption.
%
%
%
\bibliographystyle{splncs04}
\bibliography{mybibliography}

\end{document}